# STDA: Spatio-Temporal Dual-Encoder Network Incorporating Driver Attention to Predict Driver Behaviors Under Safety-Critical Scenarios

Dongyang Xu, Yiran Luo, Tianle Lu, Qingfan Wang, Qing Zhou, Bingbing Nie

*Abstract*—Accurate behavior prediction for vehicles is essential but challenging for autonomous driving. Most existing studies show satisfying performance under regular scenarios, but most neglected safety-critical scenarios. In this study, a spatio-temporal dual-encoder network named STDA for safety-critical scenarios was developed. Considering the exceptional capabilities of human drivers in terms of situational awareness and comprehending risks, driver attention was incorporated into STDA to facilitate swift identification of the critical regions, which is expected to improve both performance and interpretability. STDA contains four parts: the driver attention prediction module, which predicts driver attention; the fusion module designed to fuse the features between driver attention and raw images; the temporary encoder module used to enhance the capability to interpret dynamic scenes; and the behavior prediction module to predict the behavior. The experiment data are used to train and validate the model. The results show that STDA improves the G-mean from 0.659 to 0.719 when incorporating driver attention and adopting a temporal encoder module. In addition, extensive experimentation has been conducted to validate that the proposed module exhibits robust generalization capabilities and can be seamlessly integrated into other mainstream models.

## I. INTRODUCTION

In autonomous and semi-autonomous vehicles, the ability to understand and predict driver behavior plays a critical role in ensuring the safety and efficiency of road transportation [1, 2]. Recent advancements in machine learning and computer vision have paved the way for significant improvements in this area [3, 4]. However, the dynamic and complex nature of the driving environment introduces unique challenges, particularly in safety-critical scenarios where the timely and accurate prediction of driver actions can make the difference between safety and hazard [5, 6].

Incorporating driver attention into predictive models represents a significant departure from conventional approaches. Skilled drivers have the ability to detect and foresee potential traffic hazards rapidly. Some studies indicate that driver attention (DA) is a crucial risk indicator, proven to accurately predict driving patterns or vehicular movement, which are vital components in the decision-making process, especially under safety-critical conditions [7-10]. Moreover, both naturalistic driving assessments and controlled laboratory simulations consistently confirm the crucial role of DA in identifying objects that could lead to conflicts [11, 12].

Traditional models often depend solely on external environmental factors and historical driver data, overlooking the crucial element of where and how drivers focus their attention in varying situations [13, 14]. To bridge this gap, we introduce the STDA (Fig. 1): Spatio-Temporal dual-encoder network incorporating Driver Attention, by integrating driver attention into the driver behavior prediction model, thus offering a more holistic view of the driver's state and intentions. Furthermore, we designed a temporal encoding module to enhance the ability of the model to comprehend temporal dynamics. By merging spatial information from the driving environment with temporal patterns of driver behavior and attention, STDA seeks to offer a comprehensive framework for predicting driver actions more accurately and swiftly than before.

The contributions of this study are summarized as follows:

1) We developed a driver behavior model named STDA, tailored explicitly for safety-critical scenarios. It weaves together spatial and temporal encodings, yielding more precise predictions of driver behaviors.

2) We integrated driver attention into the STDA to better align the model with human cognitive processes and enhance its interpretability. Moreover, we explored methods for effectively fusing driver attention data with input images.

3) We conducted a comprehensive suite of experiments and ablation studies to demonstrate its superior efficacy and ability to maintain computational expeditiousness.

The structure of this paper is as follows: in Section II, we detail the previous relevant studies focused on driver attention prediction and driver behavior prediction under safety-critical scenarios. Subsequently, Section III elaborates on the formulation of the STDA with four critical parts: the driver attention prediction module, the fusion module, the temporary encoder module, and the behavior prediction module. The results and analysis of the experiments are presented in Section IV, followed by a discussion and conclusion in Section V.

This study was supported by the National Natural Science Foundation of China (52072216).

Dongyang Xu, Yiran Luo, Tianle Lu, Qingfan Wang, Qing Zhou, and Bingbing Nie are with State Key Laboratory of Intelligent Green Vehicle and Mobility, School of Vehicle and Mobility, Tsinghua University, Beijing 100084, China. (e-mail: wqf20@mails.tsinghua.edu.cn; lutl22@mails.tsinghua.edu.cn; nbb@tsinghua.edu.cn). xudy22@mails.tsinghua.edu.cn; luoyr23@mails.tsinghua.edu.cn; zhouqing@tsinghua.edu.cn;

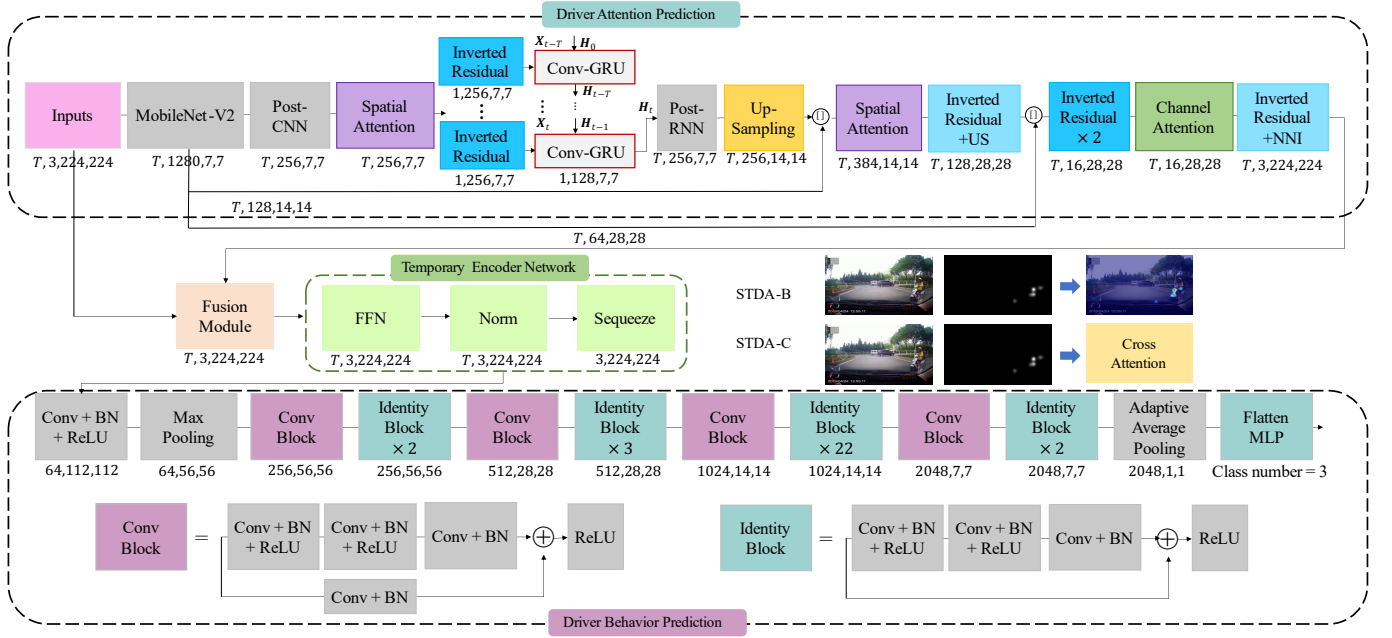

Fig. 1. Within the STDA architecture, each frame in a sequence is fed into the DA prediction models, which predict visually prominent regions that significantly stand out from their surrounding backgrounds. The feature fusion module merges the predicted data with the original first-person image streams. Then, we adopted a temporal encoding network to process the temporal features. Finally, a CNN-based deep learning model was adopted to extract high-order features, average pooling was used to obtain global features, and an MLP was utilized to predict driver behaviors.

## II. RELATED WORKS

### A. Driver Attention Prediction

The field of driver attention prediction has seen significant advancements in recent years. [15] employs U-Net as the backbone and incorporates the Swin-transformer to predict DA. [16] merges a transformer with a convolution network and then utilizes a Conv-LSTM to process the features for DA prediction. [17] utilizes a Convolutional Long Short-Term Memory (Conv-LSTM) network to capture temporal characteristics and employs a pyramid-dilated convolution approach to extract spatial attributes. It then leverages an attention mechanism to combine these temporal and spatial features, using the fused features to predict DA. Recently, inverse reinforcement learning (IRL) has represented a significant advancement in imitating driver attention, particularly in scenarios involving imminent rear-end collisions. This approach utilizes rich visual inputs, such as semantic cues, depth perception, and road lane information, extracted through pre-trained convolutional neural network (CNN)-based models. Although including detailed visual information enhances the model's performance, it also introduces a higher computational complexity than traditional end-to-end CNN architectures [18]. However, these models still encounter limitations in terms of downstream expansion.

### B. Driver Behavior Prediction

Driver behavior prediction is an essential aspect of intelligent transportation systems. Various methodologies for predicting driver behavior have been adopted, ranging from traditional statistical models to advanced machine-learning techniques. These models are broadly categorized into two distinct types: model-driven methods and data-driven methods. Each category signifies a unique approach to understanding and forecasting driving behaviors, leveraging various methodologies and theoretical foundations to enhance prediction accuracy and reliability across diverse driving scenarios. [19] develops a state space framework that integrates a Markov chain to accurately predict the likelihood of vehicles either shifting to a different lane or maintaining their current path. [20] develops mathematical models that utilize kinematic characteristics, such as velocity and steering wheel movement, to predict the likelihood of a lane change. However, model-based methods suffer from poor generalization and present difficulties in parameter calibration. For data-driven methods, [21] introduces a sophisticated hierarchical reinforcement learning framework to optimize making and executing lane change decisions. [22] leverages the RNN-LSTM architecture to create a time-series model for analyzing driving behaviors, utilizing vision-based signals to discern intentions for lane changes. However, the methods above do not consider driver attention and are not oriented towards dangerous scenarios.

## III. METHODOLOGY

In this study, we proposed a spatio-temporal dual-encoder network incorporating driver attention (STDA) for safety-critical scenarios (Fig. 1), which is aligned with the hazard perception mechanism of human drivers. STDA takes first-person image streams as input and outputs driver behaviors. Given that drivers make comprehensive decisions based on historical data. To improve the performance of the model, images from historical periods were used as inputs. Specifically, the traffic frame sequence $S_t = \{F_{t-T}, ..., F_t\}$ was sampled from the image streams, where $F_t$ represents the traffic frame to which the driver responded in safety-critical scenarios. $T$ is the length of historical frames. Each of the frames in a sequence was fed into the DA prediction models, which predict visually prominent regions that stand out from their surrounding backgrounds. The feature fusion module, which combines the predicted DA areas with the original first-person image streams, named the DA-integrating feature. A temporal encoding network processes DA-integrating features to obtain temporal features. A CNN-based deep learning model was adopted to extract high-order features, and an MLP was utilized to predict driver behaviors. Subsequent sections will introduce the driver attention prediction module, the fusion module, the temporary encoder module, and the driver behavior prediction module.

*A. Driver Attention Prediction*

Predicting where a driver is looking can help an autonomous vehicle better understand traffic situations, much like a human driver would. In the STDA model for predicting driver attention, we employed an encoder-decoder architecture. The encoder utilizes MobileNet-V2 as its backbone, owing to its low memory usage and rapid prediction capabilities. Next, we adopted Post-CNN to post-process the extracted feature $M_t \in \mathbb{R}^{T \times 1280 \times \frac{H}{32} \times \frac{W}{32}}$ to obtain $P_t \in \mathbb{R}^{T \times 256 \times \frac{H}{32} \times \frac{W}{32}}$ to reduce the amount of calculation. A spatial-attention mechanism was implemented to process $P_t$, facilitating the selective processing of visual information by prioritizing relevant regions within the visual field, thereby enhancing perceptual efficiency and accuracy in analyzing complex scenes. Formally, Conv2d projections were adopted to compute a set of queries, keys, and values (Q, K and V),

$$Q = W_q * X, K = W_k * X, V = W_v * X \quad (1)$$

where $X \in \mathbb{R}^{T \times 256 \times \frac{H}{32} \times \frac{W}{32}}$, and $*$ denotes the convolution operator. The weight matrices of $W_q, W_k$, and $W_v$ refer to the self-learned parameters of convolution operations with a stride is 1 and a filter size of 1x1. We converted the $Q, K, V \in \mathbb{R}^{T \times 256 \times \frac{H}{32} \times \frac{W}{32}}$ to $Q, K, V \in \mathbb{R}^{T \times SP \times 256}$ to match the dimensions required for the attention computation process, where $SP$ denotes the spatial dimension $\frac{H}{32} \times \frac{W}{32}$. The scaled dot product was utilized to calculate the attention weights between Q and K, to determine the spatial attention by aggregating V for each query,

$$A = \varepsilon \times softmax(\frac{QK^T}{\sqrt{D_k}})V + X' \quad (2)$$

where $D_k$ is the number of channels, values of $X'$ and $X$ are identical, $X' \in \mathbb{R}^{T \times SP \times 256}$ after reshaping, and $\varepsilon$ is a learnable parameter. We converted $A \in \mathbb{R}^{T \times SP \times 256}$ into $A \in \mathbb{R}^{T \times 256 \times \frac{H}{32} \times \frac{W}{32}}$ to meet the input requirements of subsequent modules. The processed features $A$ were passed into an inverted residual block and a Conv-GRU with 128 hidden channels and a 3x3 kernel size for sequence prediction. Enhancement of spatial-temporal feature extraction is facilitated through the integration of two critical gating mechanisms in the Conv-GRU. The complete Conv-GRU within STDA is represented as follows:

$$R_t = \Psi(BN(W_{ar} * A'_t) + BN(W_{hr} * H_{t-1}) + b_r) \quad (3)$$

$$Z_t = \Psi(BN(W_{az} * A'_t) + BN(W_{hz} * H_{t-1}) + b_z) \quad (4)$$

$$\widetilde{H}_t = \Phi \begin{pmatrix} BN(W_{ah} * A'_t) + \\ BN(W_{hh} * (R_t \odot H_{t-1})) + b_h \end{pmatrix} \quad (5)$$

$$H_t = (1 - Z_t) \odot H_{t-1} + Z_t \odot \widetilde{H}_t \quad (6)$$

where $W$ represents the weights. $\Psi$ and $\Phi$ denote the sigmoid function and the hyperbolic tangent respectively; $\odot$ denotes the Hadamard product. For the decoder, a Post-CNN was adopted to enrich the extracted feature channels, and up-sampling was utilized to align the upstream features for residual connection. Subsequently, the spatial attention mechanism was applied to enhance spatial features, followed by an inverted residual block and up-sampling to decrease the channel dimensions for improved feature representation. Two inverted residual blocks were utilized to further enrich features, thereby facilitating prediction. Additionally, another self-attention layer was added to enhance channel information. Finally, an inverted residual block was adopted to reduce the channel dimensions, and the features were upsampled to match the input image size using nearest-neighbor interpolation.

The traffic frame sequence $S_t \in \mathbb{R}^{T \times 3 \times H \times W}$ was fed to the DA prediction model, which outputs features $A_t \in \mathbb{R}^{T \times 1 \times H \times W}$ representing the driver attention at each timestep $t$, and we adopted channel extension to get $A_t \in \mathbb{R}^{T \times 3 \times H \times W}$. We passed the driver attention image stream and the original image stream into the fusion module.

*B. Fusion Module*

Two fusion approaches were designed further to explore the impact of fusion methods on model performance. The first approach, **STDA-B**, employs an image blending tactic that precisely amalgamates the original image with the driver attention heatmap on a per-pixel basis. This integration delineates areas of primary importance on the original image, enhancing the neural network's ability to focus on and assimilate pivotal visual elements. It can be calculated as:

$$F_t^f = Blend(F_t, A_{t,i}) \quad (7)$$

where $F_t^f \in \mathbb{R}^{3 \times H \times W}$ represents the fused image at each timestep $t$, and $A_{t,i} \in \mathbb{R}^{3 \times H \times W}$ is the driver attention heatmap corresponding to the $i^{th}$ frame, and the $A_{t,i}$ corresponds to the $F_t$. The fused frame sequence $S_t^f = \{F_{t-T}^f, ..., F_t^f\}$.

Another method, termed **STDA-C**, utilizes the cross-attention mechanism. STDA-C is predicated on the cross-attention mechanism and capitalizes on the synergistic integration of the original image with the driver attention heatmap. This method strategically directs focus towards regions of interest with enhanced precision by the neural network. It maintains the original image's contextual integrity and emphasizes the salient areas as delineated by the attention heatmap. The cross-attention mechanism facilitates the selective prioritization of the most pertinent features, enabling a more nuanced and compelling analysis. This approach is anticipated to enhance the neural network's performance. It can be calculated as:

$$S_t^f == LN(A_t^{mlp} + softmax(\frac{A_t^{mlp} S_t^{mlp^T}}{\sqrt{D_k}})S_t^{mlp}) \quad (8)$$

where MLP was utilized to project the images features into high dimensional space. $A_t^{mlp} = MLP(A_t)$, with $A_t^{mlp} \in \mathbb{R}^{T \times HW \times 64}$ and $S_t^{mlp} \in \mathbb{R}^{T \times HW \times 64}$, where $HW$ means $H \times W$. $LN$ denotes the Layer Normalization. To maintain consistency in the size of the fused feature output with the original, another MLP was used to back-project, ensuring the dimensions remain unchanged.

The feature fusion module, which combines the predicted DA areas with original image streams, named the DA-integrating feature $S_t^f \in \mathbb{R}^{T \times 3 \times H \times W}$.

### C. Temporary Encoder Module

In the STDA model, temporal encoding was employed to enhance the neural network's ability to recognize and interpret temporal patterns over time. The input to this module is the DA-integrating feature $S_t^f$. A feed-forward network (FFN) was used to project $S_t^f$ into a high-dimensional space, enabling the model to assimilate and integrate more comprehensive information within an intricate representational framework. We utilized Batch Normalization (BN) to standardize the processed high-dimensional features, enhancing the stability of the training process. Ultimately, the time dimension was compressed to extract the final features $S_t^{out}$. It can be calculated as:

$$S_t^{out} = Squeeze(BN(FFN(S_t^f))) \quad (9)$$

where FFN was used to project the image features into a high-dimensional space. Conversion to a high-dimensional space can improve STDA's ability to interpret time series data. Subsequently, the time dimension was compressed to convert $S_t^f \in \mathbb{R}^{T \times 3 \times H \times W}$ into $S_t^{out} \in \mathbb{R}^{3 \times H \times W}$. By reverting it to the traditional image feature dimensions, integration with existing mainstream image models is seamlessly facilitated.

The temporal encoder network enables STDA to efficiently encode and decode temporal information while preserving image resolution, significantly enhancing the capability to interpret dynamic scenes.

### D. Behavior Prediction Module

The processed DA-integrating feature $S_t^{out}$ was passed into a CNN-based network to further extract high-order features. The STDA process starts with the input being processed through a convolutional layer equipped with Batch Normalization and ReLU activation, followed by Max Pooling to reduce spatial dimensions and highlight key features. Sequentially arranged Convolutional and Identity Blocks increase the depth of network, crucial for capturing complex patterns in the data. As the architecture deepens, the number of channels increases, encoding more detailed information, while the spatial resolution decreases, focusing the network on high-level abstractions. This process is followed by Adaptive Average Pooling and a Flatten layer, culminating in a multi-layer perceptron that outputs the predicted behavior.

## IV. EXPERIMENTAL RESULTS AND ANALYSIS

### A. Datasets

In this study, the Personalized Situation Awareness of Drivers (PSAD) dataset [23] was utilized to train our model, the distribution of the dataset can be seen in Fig. 2. The PSAD dataset offers multi-modal data, including first-person driving recorder traffic frames, driver attention distribution information, and driver emergency response information. It is suitable for training driver behavior prediction models. The PSAD establishment process unfolds as follows: 2724 safety-critical scene videos were collected from driving recorders; volunteers were recruited to watch the videos in a laboratory environment; their visual gaze information was collected using eye trackers, and their emergency responses were collected using driving simulators. Due to potential distractions faced by the volunteer during the experiment, we carefully selected data from subject 101 for model training because we think subject 101 was the most careful during the experiment. Analysis reveals that, when confronted with various safety-critical scenarios, the frequencies of braking, turning right, and turning left by the volunteer were 1730, 319, and 264, respectively.

Analysis of the data composition reveals substantial non-uniformity. Applying brakes is the predominant response among drivers when encountering perilous circumstances, with only a minority opting for steering. Therefore, creating an appropriate framework to enhance model performance in the face of long-tail challenges is paramount.

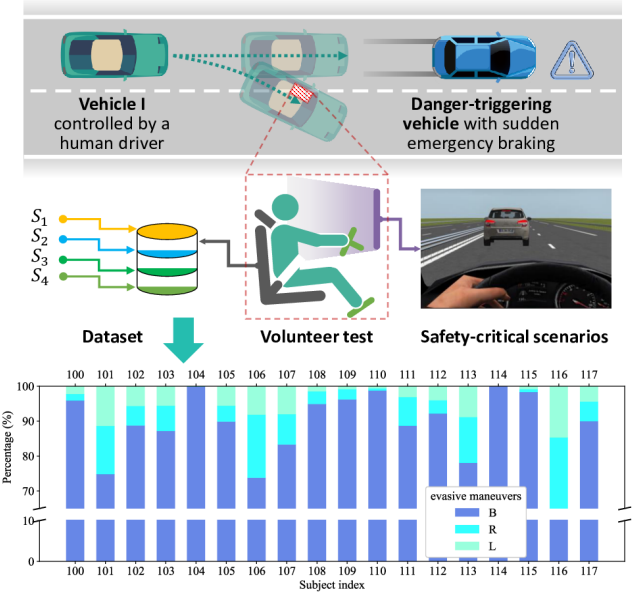

Fig. 2. The overview of dataset. Data were collected on a driving simulator by recruiting volunteers. By observing the data, we can find that the data composition reveals a substantial imbalance.

### B. Loss Function

We adopted Cost-Sensitive Learning (CSL) [28] to help the model learning. It seeks to re-balance classes by adjusting loss values for different classes during training, by assigning a higher cost to the misclassification of minority classes in imbalanced datasets, effectively shifting the classifier focus

towards these typically underrepresented classes. Specifically, it can be calculated as:

$$E_{cost} = \sum_{i=1}^{N}\sum_{j=1}^{N} C_{ij} CE(y_i j) \quad (10)$$

where $N$ is the number of classes, $C_{ij}$ is the cost of misclassifying an example from true class $i$ as class $j$. The $y_i$ is the predicted probabilities of class $i$, and $CE$ is the cross-entropy loss. The purpose of CSL is to minimize the misclassification costs ($E_{cost}$).

*C. Evaluation Metrics*

Due to the limited value of accuracy in unbalanced datasets, where correctly predicting braking can lead to misleadingly high results, this study does not employ accuracy. To undertake a comprehensive performance assessment of our proposed model, particularly consideration of data imbalance, we employed a suite of six evaluative metrics. Recall measures the model's capability to correctly identify true positive instances within the pool of actual positives. Precision gauges the accuracy with which the model classifies instances as positive.

To enhance the evaluation, four additional metrics were employed. The F1-Score, a synthesized metric, blends precision and recall. The Geometric Mean (**G-mean**) evaluates performance uniformity across different class categories, critical in the context of imbalanced data. The Index of Balanced Accuracy (**IBA**) [29], particularly relevant for our dataset, averages sensitivity and specificity, offering an equitable view of performance across various classes. Another metric is Specificity, crucial for imbalanced datasets, which assesses the model's ability to correctly classify true negatives. These additional metrics are particularly salient in our analysis, reflecting the model's adeptness at handling imbalanced class distributions.

*D. Implementation Details*

We implemented STDA within the PyTorch. For the images, the resolution of input images were cropped to 224 × 224. Following [10], we first pre-trained the DA prediction model, for training the DA model. We initialized the learning rate at 0.02 and decayed it exponentially by a factor of 0.8 after each epoch. We used stochastic gradient descent with a momentum of 0.9 for optimization. For training the whole model (STDA), we adopted Cost-sensitive Learning as loss function and used the Adam as the optimizer to train the model, the learning rate was 0.0001 and batch size was 2 per card. STDA was trained on V100 16G GPU. As for the throughout, we configured the batch size to 256 and conducted assessments on an A800 80G GPU.

TABLE I. COMPARING THE RESULTS BETWEEN OUR MODEL AND EXISTING MAINSTREAM MODELS, G-MEAN AND IBA ARE THE MAIN METRICS, RESULTS IS THE AVERAGES OF EACH CLASS.

| Model | Params | FLOPs | Throughout (image/s) | Precision | Recall | Specific | F1-score | Geo-mean | Index of Balanced Accuracy (IBA) |
|---|---|---|---|---|---|---|---|---|---|
| Machine Learning | | | | | | | | | |
| DT | — | — | — | 0.567 | 0.575 | 0.414 | 0.571 | 0.402 | 0.167 |
| Adaboost | — | — | — | 0.542 | 0.715 | 0.280 | 0.611 | 0.104 | 0.014 |
| RF | — | — | — | 0.569 | 0.706 | 0.300 | 0.615 | 0.195 | 0.042 |
| Resnet | | | | | | | | | |
| Resnet-18 [24] | 11.7M | 1.83G | 1425.3 | 0.738 | 0.755 | 0.518 | 0.742 | 0.580 | 0.345 |
| Resnet-50 [24] | 25.6M | 4.13G | 1085.1 | 0.716 | 0.692 | 0.570 | 0.701 | 0.602 | 0.368 |
| Resnet-101 [24] | 44.6M | 7.87G | 975.2 | 0.731 | 0.673 | 0.660 | 0.693 | 0.659 | 0.435 |
| Vision-Transformer | | | | | | | | | |
| ViT-T/16 [25] | 5.72M | 0.92G | 1015.6 | 0.685 | 0.572 | 0.555 | 0.596 | 0.529 | 0.291 |
| ViT-S/16 [25] | 22.05M | 3.22G | 491.2 | 0.716 | 0.709 | 0.564 | 0.712 | 0.606 | 0.372 |
| ViT-B/16 [25] | 86.57M | 12.02G | 319.5 | 0.721 | 0.731 | 0.528 | 0.724 | 0.582 | 0.346 |
| ViT-L/16 [25] | 304.33M | 41.82G | 188.5 | 0.714 | 0.718 | 0.561 | 0.715 | 0.601 | 0.370 |
| DistillableViT (Do not Pretrain) | | | | | | | | | |
| DistillableViT/L2 | 13.6M | 2.72G | 1086.8 | 0.640 | 0.650 | 0.459 | 0.643 | 0.475 | 0.238 |
| DistillableViT/L6 | 38.8M | 7.84G | 771.4 | 0.637 | 0.697 | 0.384 | 0.662 | 0.403 | 0.173 |
| DistillableViT/L12 | 76.59M | 15.53G | 537.9 | 0.635 | 0.719 | 0.326 | 0.662 | 0.318 | 0.112 |
| Swin-Transformer | | | | | | | | | |
| Swin-T [26] | 28.29M | 4.38G | 644.8 | 0.737 | 0.752 | 0.544 | 0.743 | 0.602 | 0.371 |
| Swin-S [26] | 49.6M | 8.56G | 426.1 | 0.730 | 0.751 | 0.518 | 0.738 | 0.578 | 0.343 |
| Swin-B [26] | 87.77M | 15.19G | 352.1 | 0.732 | 0.715 | 0.599 | 0.722 | 0.636 | 0.409 |
| Swin-L [26] | 196.53M | 34.12G | 260.9 | 0.758 | **0.777** | 0.556 | **0.763** | 0.618 | 0.392 |
| ConvNeXt | | | | | | | | | |
| ConvNeXt-T [27] | 28.59M | 4.48G | 841.8 | 0.706 | 0.592 | 0.646 | 0.624 | 0.610 | 0.372 |
| ConvNeXt-S [27] | 50.22M | 8.72G | 607.9 | 0.756 | 0.771 | 0.555 | 0.761 | 0.619 | 0.392 |
| ConvNeXt-B [27] | 88.59M | 15.41G | 499.1 | 0.733 | 0.746 | 0.534 | 0.738 | 0.593 | 0.359 |
| ConvNeXt-L [27] | 197.77M | 34.44G | 300.9 | 0.743 | 0.762 | 0.537 | 0.750 | 0.599 | 0.367 |
| STDA (Ours) | | | | | | | | | |
| STDA-B | 54.92M | 8.65G | 757.6 | **0.769** | 0.692 | **0.749** | 0.714 | **0.719** | **0.513** |
| STDA-C | 54.92M | 8.70G | 732.3 | 0.734 | 0.646 | 0.696 | 0.674 | 0.667 | 0.443 |

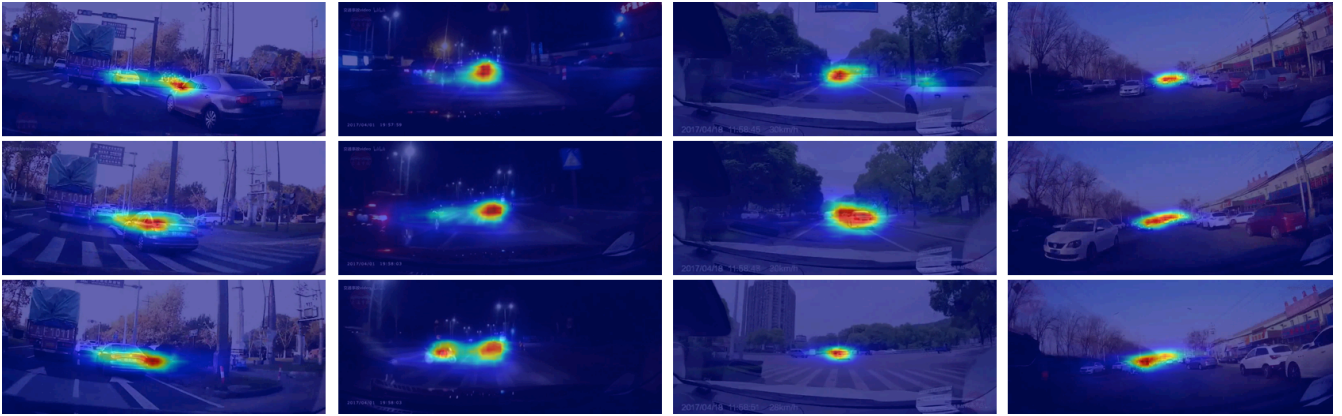

Fig. 3. The impact of DA module on model prediction. Each row is a scene, and each column is the evolution of any scene over time. The outcomes predicted by the driver attention (DA) module, highlighting the areas that require the driver's heightened vigilance.

*E. Comparison with Existing Models*

Analyzing TABLE I, regarding computational efficiency, the FLOPs for the STDA variants are 8.65G and 8.70G, respectively. This places them in the higher echelon of computational intensity. Despite this, the throughput efficiency, measured at 757.6 and 732.3 images per second for STDA-B and STDA-C, respectively, signifies a commendable processing velocity compared to existing models. Such metrics underscore the model's high computational efficiency, which is crucial in safety-critical scenarios where the window of time to situations is extremely limited.

The STDA-B model, when compared to other mainstream models. The specificity of the STDA-B model is noteworthy, reflecting its effectiveness in accurately identifying negative instances and thus reducing false positives, a vital feature when handling imbalanced datasets. This is further supported by its F1-Score, which, while not at the very top, signifies a well-rounded balance between precision and recall. Considering the G-mean and IBA scores, STDA-B stands out in maintaining performance consistency across classes, which is beneficial when dealing with varied class distributions, especially in imbalanced contexts. STDA-B prioritizes balanced performance across classes rather than achieving high accuracy on the majority class at the expense of minority classes. In summary, the experimental results demonstrate that STDA is effective for imbalanced datasets, ensuring that it does not disproportionately favor the majority class. STDA is suitable for scenarios like safety-critical conditions where hazardous but rare events form a long-tail distribution.

The underwhelming performance of knowledge distillation models might stem from the distillation process's complexities and mismatches between teacher and student models. Additionally, imblanced datasets may lead to overfitting on dominant classes while overlooking rarer ones. Moreover, we failed to pretrain it may also be a potential reason for its poor effect. Addressing data imbalances and incorporating pretraining stages could further improve model robustness and performance.

It is observed that STDA-B outperforms STDA-C in prediction performance. The superior efficacy of image blending compared to cross-attention-based fusion can be attributed to its direct and uniform integration of features, which can preserve spatial coherency and reduce feature distraction. In contrast to cross-attention mechanisms that selectively emphasize features and may inadvertently skew feature representation towards dominant patterns, blending ensures a more holistic and balanced integration of visual data, potentially yielding a richer and more accurate feature landscape for analysis.

*F. Ablation Studies*

Ablation experiments on STDA were conducted to determine the utility of the designed modules for the model. As for removing the DA module and temporary module, we initialized the DA module to ensure the proper functioning of the fusion module. Results are presented in TABLE II. For STDA-B, incorporating the DA module resulted in a 4.5% increase in G-Mean, highlighting significant improvement in handling imbalanced datasets. The temporary encoder module alone did not significantly affect the F1-score or Specificity but led to a modest increase in G-Mean. Employing both modules yielded appreciable improvements across all metrics, especially in G-Mean and IBA, with increases of 8.9% and 18%, respectively. For STDA-C, the enhancements were even more pronounced. The combination DA module with temporary module significantly boosted G-Mean by 28.3%. These findings highlight the synergistic impact of the DA and temporary encoder modules on enhancing STDA's performance.

The DA module improves the model's comprehension of its surroundings by emphasizing the areas requiring the most attention, enhancing its robustness and generalization ability. For instance, within the safety-critical scenarios depicted in Fig. 3, When the car is about to change lanes, DA module will immediately pay attention to the surrounding vehicles interacting, providing a prior knowledge to the behavior prediction model. This augmentation enables timely and informed decision-making in complex traffic conditions. The temporary encoder module enhances STDA's temporal feature encoding capabilities, enriching its feature space for more nuanced data interpretations. Combining these modules performs well across individual classes and maintains uniform performance across the dataset, achieving the best performance, as evidenced by the significant increases in G-Mean and IBA.

TABLE II. ABLATION STUDIES DEMONSTRATE THAT INCORPORATING DRIVER ATTENTION AND TEMPORARY ENCODING SIGNIFICANTLY ENHANCES PERFORMANCE

| Model | DA module | Temporary module | Precision | Recall | F1-score | Specific | IBA | Geo-mean | Geo-mean Increase |
|---|---|---|---|---|---|---|---|---|---|
| STDA-B | — | — | 0.731 | 0.673 | 0.693 | 0.660 | 0.435 | 0.659 | — |
|  | √ | — | 0.757 | **0.725** | **0.737** | 0.671 | 0.477 | 0.689 | +4.5% |
|  | — | √ | 0.748 | 0.712 | 0.724 | 0.660 | 0.458 | 0.674 | +2.2% |
|  | √ | √ | **0.769** | 0.692 | 0.714 | **0.749** | **0.513** | **0.719** | **+8.9%** |
| STDA-C | — | — | 0.667 | **0.661** | 0.664 | 0.484 | 0.276 | 0.520 | — |
|  | √ | — | 0.706 | 0.647 | 0.669 | 0.616 | 0.387 | 0.621 | +19.4% |
|  | — | √ | 0.694 | 0.562 | 0.592 | 0.635 | 0.335 | 0.570 | +9.7% |
|  | √ | √ | **0.734** | 0.646 | **0.674** | **0.696** | **0.443** | **0.667** | **+28.3%** |

TABLE III. ABLATION STUDIES SHOW THAT INTEGRATING DRIVER ATTENTION AND TEMPORARY ENCODING INTO EXISTING MODELS SIGNIFICANTLY ENHANCES PERFORMANCE

| Model | DA module | Temporary module | IBA | Geo-mean | Geo-mean Increase | Model | DA module | Temporary module | IBA | Geo-mean | Geo-mean Increase |
|---|---|---|---|---|---|---|---|---|---|---|---|
| Resnet18 | — | — | 0.345 | 0.580 | — | Resnet50 | — | — | 0.368 | 0.602 | — |
|  | √ | — | 0.422 | 0.648 | +11.7% |  | √ | — | 0.427 | 0.649 | +7.8% |
|  | — | √ | 0.398 | 0.625 | +7.8% |  | — | √ | 0.431 | 0.654 | +8.6% |
|  | √ | √ | 0.430 | 0.652 | **+12.4%** |  | √ | √ | 0.428 | 0.656 | **+8.9%** |
| Resnet101 | — | — | 0.435 | 0.659 | — | ViT-T | — | — | 0.291 | 0.529 | — |
|  | √ | — | 0.477 | 0.689 | +4.5% |  | √ | — | 0.387 | 0.619 | **+17.0%** |
|  | — | √ | 0.458 | 0.674 | +2.2% |  | — | √ | 0.367 | 0.604 | +14.2% |
|  | √ | √ | 0.514 | 0.719 | **+8.9%** |  | √ | √ | 0.376 | 0.607 | +14.7% |
| ViT-S | — | — | 0.372 | 0.606 | — | ViT-B | — | — | 0.346 | 0.582 | — |
|  | √ | — | 0.421 | 0.646 | +6.7% |  | √ | — | 0.414 | 0.639 | +9.9% |
|  | — | √ | 0.383 | 0.619 | +2.3% |  | — | √ | 0.430 | 0.655 | +12.7% |
|  | √ | √ | 0.426 | 0.650 | **+7.4%** |  | √ | √ | 0.434 | 0.655 | **+12.7%** |
| ViT-L | — | — | 0.370 | 0.601 | — | Swin-T | — | — | 0.371 | 0.602 | — |
|  | √ | — | 0.382 | 0.612 | +1.9% |  | √ | — | 0.393 | 0.621 | +3.1% |
|  | — | √ | 0.422 | 0.647 | +7.6% |  | — | √ | 0.408 | 0.637 | +5.7% |
|  | √ | √ | 0.452 | 0.676 | **+12.6%** |  | √ | √ | 0.428 | 0.654 | **+8.6%** |
| Swin-S | — | — | 0.343 | 0.578 | — | Swin-B | — | — | 0.409 | 0.636 | — |
|  | √ | — | 0.404 | 0.631 | +9.1% |  | √ | — | 0.459 | 0.675 | +6.2% |
|  | — | √ | 0.381 | 0.612 | +6.0% |  | — | √ | 0.474 | 0.690 | +8.5% |
|  | √ | √ | 0.413 | 0.638 | **+10.4%** |  | √ | √ | 0.499 | 0.710 | **+11.7%** |
| Swin-L | — | — | 0.392 | 0.618 | — | Distillable ViT/L2 | — | — | 0.238 | 0.475 | — |
|  | √ | — | 0.390 | 0.619 | +0.1% |  | √ | — | 0.313 | 0.555 | +16.9% |
|  | — | √ | 0.422 | 0.647 | +4.5% |  | — | √ | 0.248 | 0.488 | +2.7% |
|  | √ | √ | 0.452 | 0.676 | **+9.3%** |  | √ | √ | 0.347 | 0.584 | **+23.1%** |
| Distillable ViT/L6 | — | — | 0.173 | 0.403 | — | Distillable ViT/L12 | — | — | 0.112 | 0.318 | — |
|  | √ | — | 0.351 | 0.590 | **+46.3%** |  | √ | — | 0.202 | 0.442 | **+39.0%** |
|  | — | √ | 0.171 | 0.403 | +0.9% |  | — | √ | 0.113 | 0.328 | +3.1% |
|  | √ | √ | 0.310 | 0.554 | +37.5% |  | √ | √ | 0.202 | 0.442 | +38.8% |
| ConvNeXt-T | — | — | 0.372 | 0.610 | — | ConvNeXt-S | — | — | 0.392 | 0.619 | — |
|  | √ | — | 0.434 | 0.656 | +7.6% |  | √ | — | 0.466 | 0.680 | +9.8% |
|  | — | √ | 0.438 | 0.657 | +7.7% |  | — | √ | 0.409 | 0.634 | +2.4% |
|  | √ | √ | 0.444 | 0.662 | **+8.6%** |  | √ | √ | 0.496 | 0.703 | **+13.5%** |
| ConvNeXt-B | — | — | 0.359 | 0.593 | — | ConvNeXt-L | — | — | 0.367 | 0.599 | — |
|  | √ | — | 0.410 | 0.633 | +6.8% |  | √ | — | 0.420 | 0.641 | **+7.1%** |
|  | — | √ | 0.369 | 0.602 | +1.6% |  | — | √ | 0.356 | 0.588 | -1.8% |
|  | √ | √ | 0.415 | 0.638 | **+7.5%** |  | √ | √ | 0.406 | 0.632 | +5.6% |

To further investigate the universality of the designed module, ablation experiments were conducted on all existing mainstream image models. The ablation study detailed in TABLE III thoroughly analyzes how the inclusion of DA and temporal modules affects performance across various models. The inclusion of the DA and temporal modules leads to performance improvements in nearly all models tested, demonstrating their effectiveness as generalizable components across diverse image model. For instance, the addition of the DA module has notably enhanced the G-mean in models like Resnet, ViT, and Swin Transformer variants, with a significant improvement of 46.3% observed in the Distillable-ViT/L6 model. From CNNs such as Resnet to Vision Transformers and their refined variants, the consistent results across these architectures confirm that DA and temporal encoder modules are beneficial for improving model performance and can be effectively used as general-purpose enhancements in neural networks.

It is important to highlight that the G-mean of ConvNext-L decreased by 1.8% following the incorporation of the temporal encoder. This decline may be attributed to the model's considerable size and the processing of the time dimension, potentially causing overfitting. This issue merits further investigation and discussion.

## V. CONCLISION

This study introduces a novel approach: the Spatio-Temporal dual-encoder incorporating Driver Attention, designed for predicting driver behavior in safety-critical scenarios. By integrating spatial and temporal data along with a focus on driver attention, the STDA enhances the G-mean and IBA metrics in predicting driver behaviors. Two fusion module, STDA-B and STDA-C are proved as effective approaches to utilize driver attention especially STDA-B which employs an image blending tactic, enhancing model's performance. The inclusion of driver attention not only aligns the model with human cognitive processes but also significantly boosts its interpretability. The STDA model maintains computational efficiency through detailed experimentation while delivering state-of-the-art performance and demonstrating robust generalization capabilities. These attributes underscore STDA's potential for seamless integration into mainstream models, broadening its utility. Future work will expand STDA's applicability to complex driving scenarios and improve its real-time processing capabilities. Additionally, field testing will be essential to move the algorithm from theoretical validation to practical implementation.